\documentclass[runningheads]{llncs}

\usepackage{eccv}

\usepackage{eccvabbrv}

\usepackage{graphicx}
\usepackage{booktabs}
\usepackage{multirow}
\usepackage{tabularx}

\usepackage[accsupp]{axessibility}  

\usepackage{hyperref}

\usepackage{orcidlink}

\begin{document}

\title{InnoText: A Unified Model for Visual Text Generation and Editing} 

\titlerunning{InnoText: A Unified Model for Visual Text Generation and Editing}

\author{Haowei Liu\inst{1,2}$^*$$^\S$\orcidlink{0009-0003-6300-5247} \and
Runze He\inst{3}$^*$\orcidlink{0009-0009-7917-7223} \and
Jian Lu\inst{4}$^*$\orcidlink{0009-0009-9872-5336} \and
Ao Ma\inst{2}$^\dagger$$^\ddagger$\orcidlink{0009-0000-1967-8020}\and
Run Ling\inst{2}\orcidlink{0009-0007-8616-8582}\and
Ke Cao\inst{2}\orcidlink{0009-0000-4421-0817}\and
Jiasong Feng\inst{5}\orcidlink{0009-0002-9542-3382}\and
Wei Feng\inst{2}\orcidlink{0009-0005-8890-4956}\and
Shuo Lu\inst{6}\orcidlink{0009-0000-7547-3169}\and
Yexing Xu\inst{2}\orcidlink{0009-0009-9688-8318}\and
Yun Wang\inst{7}\orcidlink{0000-0001-8384-6981}\and
Jing Wang\inst{1}\orcidlink{0009-0001-0392-4976}\and
Zhanjie Zhang\inst{8}$^\dagger$\orcidlink{0000-0002-8966-1328}}

\authorrunning{H.Liu et al.}

\institute{Shenzhen Campus of Sun Yat-Sen University \and
JD.com, Inc. \and
University of Chinese Academy of Sciences \and
Chongqing University of Post and Telecommunications \and
Beijing University of Technology \and
Institute of Automation, Chinese Academy of Sciences\and
City University of Hong Kong\and
Zhejiang University\\
\email{liu\_haow@163.com}\\
}



{
    \renewcommand{\thefootnote}{} 
    \footnotetext{$^*$Equal contribution. $^\dagger$Corresponding author. $^\ddagger$Project leader. $^\S$Conducted during internship.}
}

\maketitle

\begin{figure}[h!]
    \centering
    \includegraphics[width=\textwidth]{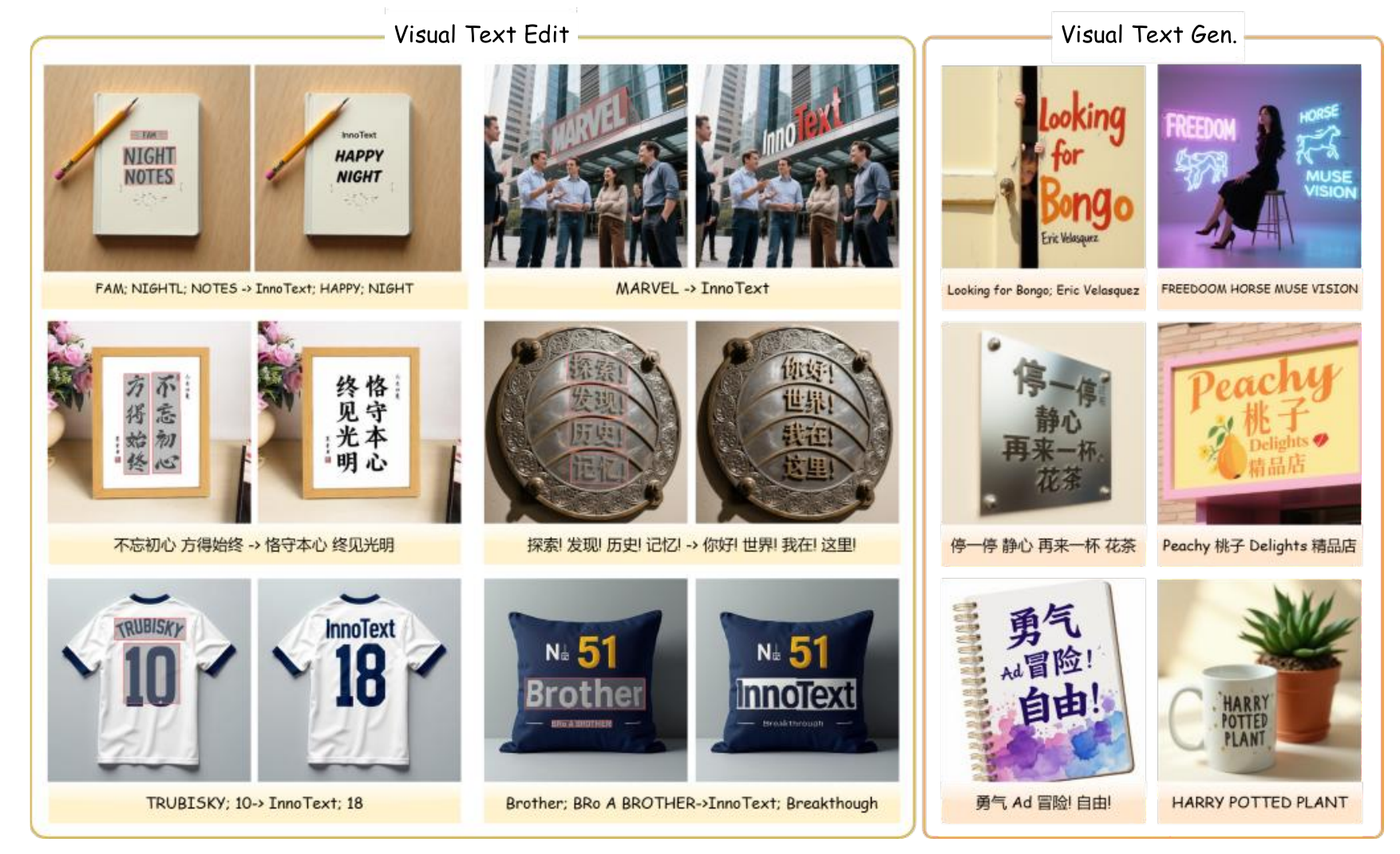}
    \caption{Visual examples generated by our method for visual text generation and editing.}
    \label{fig:fig1}
\end{figure}

\begin{abstract}
Diffusion models have recently achieved remarkable success in high-fidelity image synthesis, yet their application to visual text generation and editing remains relatively underexplored. Unlike general image generation, visual text tasks demand precise structural regularity and legibility, which may pose additional challenges for small-scale text and non-Latin scripts such as Chinese. Existing UNet-based models often struggle to produce clear and coherent text, while DiT-based models, though more expressive, are typically limited to a single task, which may lead to redundant training pipelines, inconsistent visual styles, and reduced cross-task generalization. To address these challenges, we propose InnoText, a unified DiT-based framework capable of performing both text generation and editing within a single model. We introduce a Font Size-Aware Modulation (FSAM) module to enhance representations across font scales, a Small-Character Aware Augmentation strategy to improve fine-grained fidelity, and a Task-Specific Region Weighted Loss for adaptive optimization. To support training and evaluation, we also construct a high-quality bilingual (English-Chinese) visual text dataset covering diverse fonts, sizes, and backgrounds. Experimental results demonstrate that our method achieves superior generation accuracy and editing quality, producing visually appealing and realistic text images.

\end{abstract}    
\section{Introduction}
\label{sec:intro}

\begin{figure}[b]
\centering
\includegraphics[width=\columnwidth]{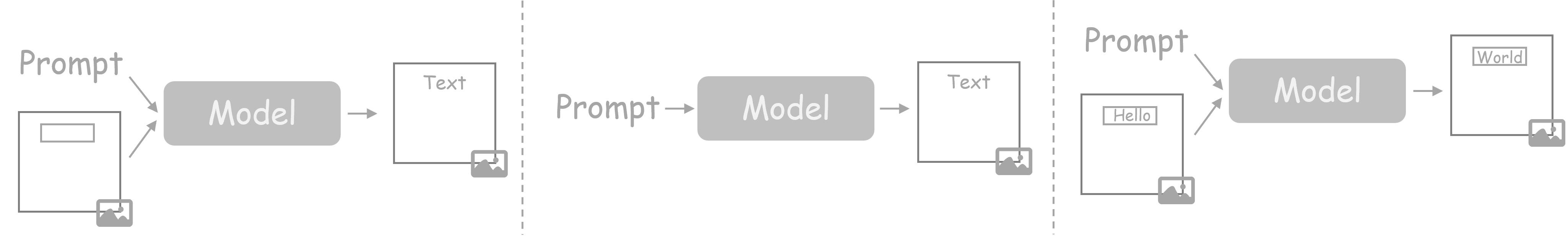}
\caption{(Left) Visual text generation in previous unified models~\cite{wang2025dreamtext,zhao2024udifftext}. (Middle) Visual text generation in AnyText~\cite{tuo2023anytext} and ours. (Right) Visual text editing task.}
\label{task_define}
\end{figure}

\begin{figure}[!t]
\centering
\includegraphics[width=\columnwidth]{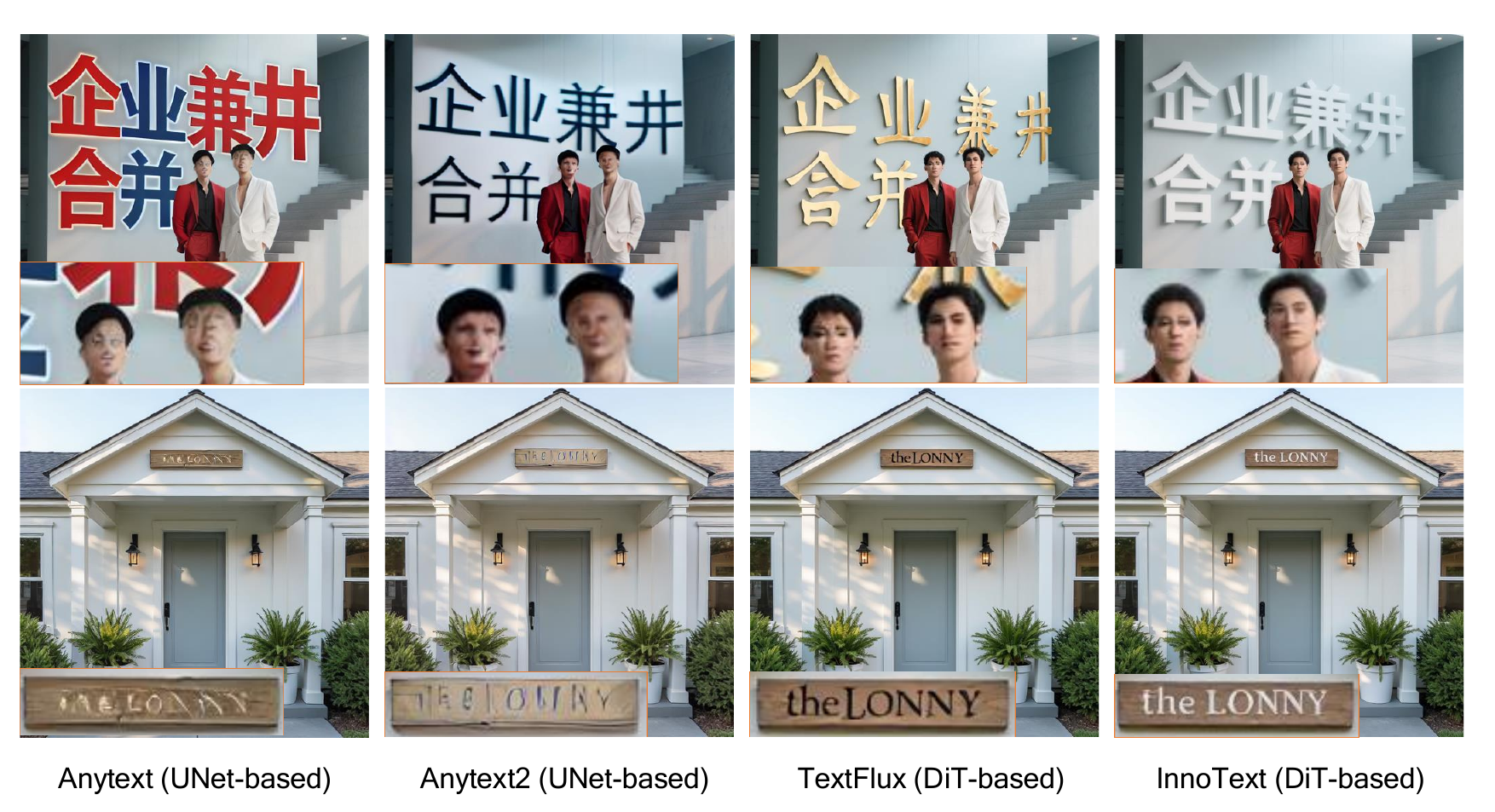}
\caption{Visual text images generated by diffusion models based on UNet and DiT, respectively.}
\label{small_font}
\vspace{-10pt}
\end{figure}

Recent advances in diffusion models~\cite{rombach2022high, esser2024scaling, flux2024} have demonstrated profound capabilities in image generation~\cite{tan2024ominicontrol, xu2026design, zhang2025u}. However, extending these successes to visual text generation and editing remains a formidable challenge. Visual text generation aims to synthesize coherent text within images directly from prompts, a task we formulate as a simultaneous text-to-image (T2I) synthesis problem~\cite{tuo2023anytext}, as shown in Fig.~\ref{task_define}. This paradigm contrasts with conventional unified models~\cite{wang2025dreamtext, zhao2024udifftext} that treat generation as a text-guided image-to-image (TI2I) insertion task. While visual text editing modifies content while preserving visual context. However, prevailing UNet-based methods~\cite{tuo2023anytext, zeng2024textctrl} often suffer from structural distortions and localized artifacts (see Fig.~\ref{small_font}). Although emerging Diffusion Transformer (DiT) architectures~\cite{peebles2023scalable, xie2025textflux} offer superior fine-grained modeling, most existing frameworks are restricted to single-task designs. Such isolation prevents the exploitation of cross-task synergies and necessitates redundant optimization. Furthermore, the inherent complexity of non-Latin scripts like Chinese, characterized by intricate glyph structures, poses significant hurdles for current models primarily optimized for English. This difficulty is exacerbated in micro-scale text scenarios where existing methods fail to maintain legibility. Moreover, the scarcity of high-resolution Chinese datasets~\cite{tuo2023anytext} further constrains model generalization to diverse real-world conditions.

To address these systemic limitations, we propose InnoText, a unified DiT-based framework that reformulates visual text tasks through in-context learning. By designing specialized input paradigm, InnoText achieves architectural unification, enabling seamless switching between generation and editing within a single model. To mitigate the scale-sensitivity issue, we introduce the Font Size-Aware Modulation (FSAM) module. FSAM dynamically re-calibrates the latent feature space via a spatial size-map, ensuring scale-invariant fidelity across diverse font dimensions. Furthermore, we propose a Small-Character Aware Augmentation (SCAA) strategy, acting as a foveal attention mechanism during training to distill sub-pixel structural priors of intricate glyphs. Finally, a Task-Specific Region Weighted Loss is implemented to balance global coherence in generation with localized precision in editing. Our approach not only establishes a versatile benchmark for bilingual visual text tasks but also provides a robust solution for preserving structural integrity in complex, multi-scale real-world scenarios.

In summary, the main contributions of InnoText are as follows:

\begin{itemize}
    \item We propose InnoText, a novel unified framework based on the DiT architecture. By leveraging in-context learning to design specialized input patterns, InnoText seamlessly integrates visual text generation and editing into a single model, offering superior flexibility and broader applicability in real-world scenarios.
    \item We introduce a Font Size-Aware Modulation (FSAM) module that leverages a size map to adaptively enhance text regions of different scales. In addition, we propose a Small-Character Aware Augmentation and a Task-Specific Region Weighted Loss to improve generation quality.
    \item We construct a high-quality visual text dataset with diverse font styles, sizes, and background complexities. Extensive experiments demonstrate the superiority of our model over existing baselines in both visual text generation and editing tasks.
\end{itemize}

\section{Related Work}
\label{sec:formatting}

\subsection{Text to Image Generation}
With the rise of diffusion models, researchers have begun exploring their potential in text-to-image generation~\cite{podell2023sdxl,esser2024scaling,flux2024}. The introduction of latent diffusion models~\cite{rombach2022high} marked a key milestone by performing diffusion in latent space, significantly reducing computational cost while preserving visual detail. While early models like Stable Diffusion~\cite{rombach2022high,podell2023sdxl} achieved strong performance in general image synthesis, recent transformer-based architectures, such as Diffusion Transformer~\cite{peebles2023scalable}, have further improved generation quality. Notable examples like SD3~\cite{esser2024scaling} and Flux~\cite{flux2024} leverage stronger architectures and techniques such as rectified flow~\cite{liu2022flow} to enhance image generation quality and efficiency. These advances have broadened the applicability of DiT-based models to downstream tasks including customized image generation~\cite{cao2026relactrl, qin2026innoads}, visual editing~\cite{zhang2025context, song2025insert}, and visual text synthesis~\cite{lu2025easytext,xie2025textflux}.

\subsection{Controllable Diffusion Transformers}
Traditional UNet-based generation methods, such as ControlNet~\cite{zhang2023adding}, IP-Adapter~\cite{ye2023ip}, and T2I-Adapter~\cite{mou2024t2i}, enable downstream applications by allowing the model to take reference images (e.g. depth maps, subject images) as conditional input~\cite{ling2026mofu,qin2026innoads}. Recently, DiT-based conditional generation~\cite{tan2024ominicontrol,zhang2025easycontrol, li2025ic, song2025insert, wang2026wisa, ma2025lay2story} has also seen significant progress. These approaches typically encode visual conditions into tokens, which are then integrated with text tokens using the existing multi-modal attention mechanisms within the DiT architecture. This design allows for effective fusion of image-based conditions without requiring substantial architectural modifications.

\subsection{Visual Text Generation and Editing}
With the rapid development of diffusion models, generating high-quality and semantically accurate visual text images has become a key focus of research. Although recent work~\cite{chen2023textdiffuser,yang2023glyphcontrol,chen2024textdiffuser, zhu2024visual} predominantly focused on synthesizing English-language text, recent studies, such as AnyText~\cite{tuo2023anytext} and GlyphByT5~\cite{liu2024glyph}, have begun exploring the challenges of generating non-Latin character text, such as Chinese. Due to the limited language understanding capabilities of standard diffusion models for non-English languages, some approaches~\cite{liu2024glyph,gong2025seedream,ma2024chargen, zhao2024udifftext} have proposed improvements to the text embedding process, for example, by leveraging large language models to enhance the semantic representation of multilingual prompts. Other methods~\cite{tuo2023anytext,tuo2024anytext2,jiang2025controltext,zhang2024control} introduce glyph images as additional visual conditions, rendering the prompt text into font-based visual representations that are fed into the model to guide more accurate visual text generation.

To improve generation quality, recent state-of-the-art methods~\cite{wang2025reptext,lu2025easytext,lan2025flux,xie2025textflux,du2025textcrafter} increasingly adopt the DiT-based architecture as the backbone. However, most existing approaches are designed for either visual text editing~\cite{xie2025textflux,lan2025flux} or generation~\cite{wang2025reptext,lu2025easytext,du2025textcrafter}, and few are capable of jointly handling both tasks within a unified framework. To address this gap, we propose a DiT-based model specifically designed for visual text editing and generation, enabling a more flexible and accurate solution for real-world multilingual text synthesis.

\section{Datasets}

\subsection{Overview}

To overcome the scarcity of high-fidelity Chinese visual text data, we meticulously curated InnoText-30K, a bilingual dataset (20K Chinese, 10K English) designed for unified text tasks. Our primary contribution lies in bridging the quality gap between Chinese and English datasets through cross-lingual semantic mapping: we extended the English Lex-10K~\cite{zhao2025lex} into a high-quality Chinese counterpart of 10K samples using advanced generative pipelines, ensuring structural and aesthetic parity across languages. This is further augmented by a strategically selected 10K-sample subset from Anyword-3M~\cite{tuo2023anytext}, integrating diverse real-world and synthetic scenarios to enhance environmental robustness.

\begin{figure}[t]
\centering
\includegraphics[width=\linewidth]{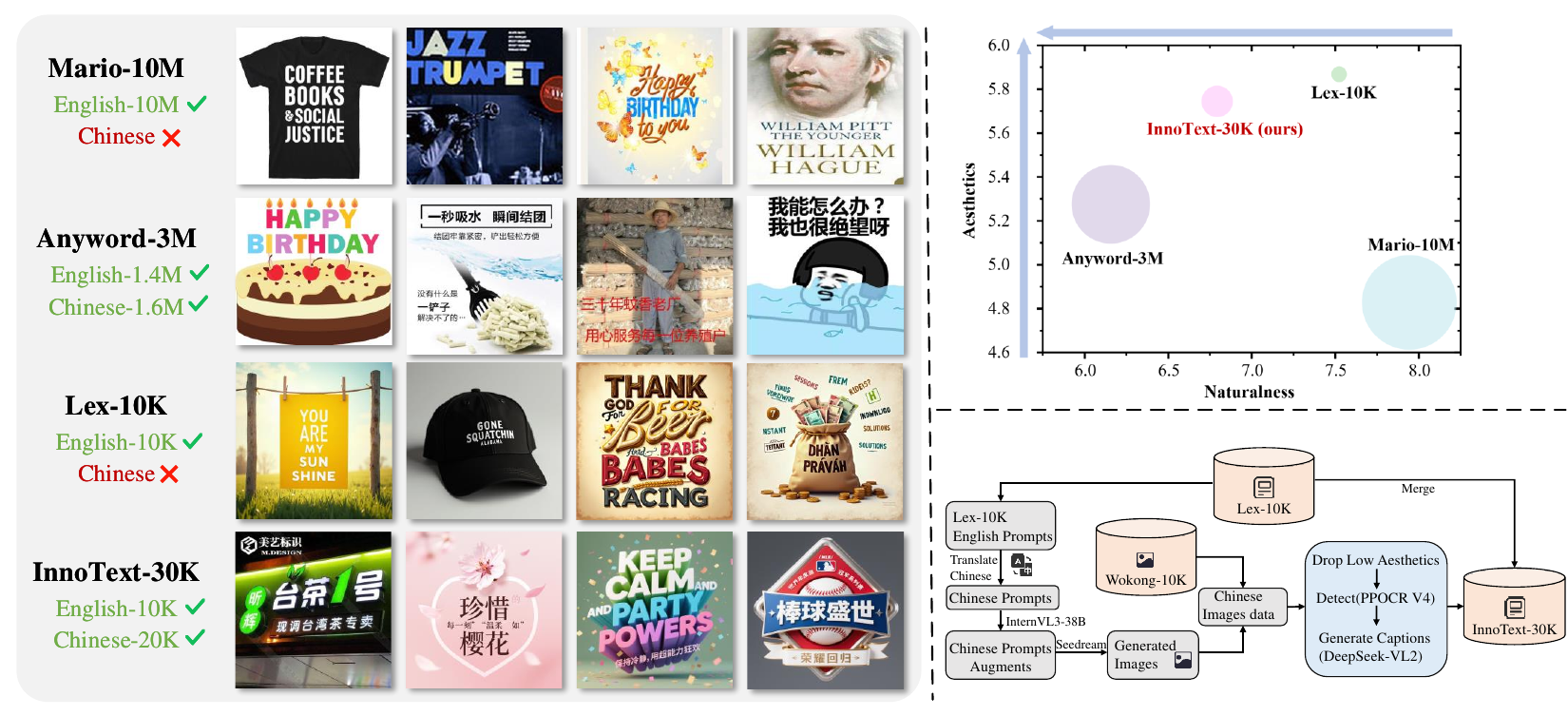}
\caption{\textbf{Datasets overview.} (Left) Representative cases from multiple datasets.  (Top right) Comparison of aesthetics and naturalness across different datasets. The aesthetic scores are obtained using the Neural Image Assessment (NIMA) model~\cite{talebi2018nima}, while the naturalness scores are evaluated using the Natural Image Quality Evaluator (NIQE) metric~\cite{mittal2012making}. (Bottom right) Dataset processing pipeline.}
\label{fig:dataset}
\end{figure}

As illustrated in Fig.~\ref{fig:dataset}, InnoText-30K achieves a superior balance between visual aesthetics (ranking second only to Lex-10K) and perceptual naturalness (surpassing Lex-10K and matching Anyword-3M). These results validate that our curated data provides a high-quality, distributionally consistent foundational corpus for fine-tuning text-centric DiT models.

\subsection{Pipeline}

The InnoText-30K construction pipeline, illustrated in Fig.~\ref{fig:dataset}, prioritizes cross-lingual semantic fidelity and visual refinement. For the synthetic component, we initiate a knowledge transfer process by translating Lex-10K~\cite{zhao2025lex} captions into Chinese, followed by a semantic expansion phase using Intern-VL3~\cite{zhu2025internvl3}. This stage leverages context-aware keyword substitution and prompt enrichment to synthesize expressive, high-entropy textual descriptors. These refined prompts are then fed into the Seedream~\cite{gong2025seedream} engine to generate Chinese image–text pairs, effectively bridging the representation gap in existing bilingual datasets.

To ensure rigorous quality control, the synthetic data is integrated with an Anyword-3M~\cite{tuo2023anytext} subset and subjected to a multi-stage curation bottleneck. First, an aesthetic scoring filter prunes samples with suboptimal composition. Subsequently, we execute precision OCR re-annotation using PPOCR-v4 to rectify spatial-textual misalignments. Finally, DeepSeek-VL2~\cite{wu2024deepseek} is employed for vision-language recalibration, generating dense captions that synchronize visual content with OCR outputs for enhanced semantic grounding.

The resulting Chinese corpus is fused with Lex-10K to form our final dataset. For robust evaluation, we establish InnoText-Bench, a diagnostic suite of 1,500 images sampled across the three constituent subsets, ensuring a balanced assessment of model performance across diverse linguistic and aesthetic distributions.

\section{Method}

\subsection{Overview}
Our framework is based on the Flux.1 Fill dev~\cite{flux2024}, a powerful inpainting model derived from the Flux backbone. While retaining its strong generative capacity, we adapt this model to support both generation and editing tasks by carefully designing the input masking strategy.

For editing tasks, we define three input components: the masked image $I_{masked}$, the glyph image $I_{glyph}$, and the corresponding mask $M$. These are combined to form the model input in the spatial dimension as follows:
\begin{equation}
I_{\text{input}} = \text{Concat}([I_{\text{glyph}}, I_{\text{masked}}], \text{dim}=0)
\end{equation}
\begin{equation}
I_{\text{mask}} = \text{Concat}([\textbf{0}, M], \text{dim}=0)
\end{equation}
For generation tasks, the model input is constructed by concatenating the glyph image with a fully black image (i.e., an empty canvas) along the spatial dimension:
\begin{equation}
I_{\text{input}} = \text{Concat}([I_{\text{glyph}}, \textbf{0}], \text{dim}=0)
\end{equation}
\begin{equation}
I_{\text{mask}} = \text{Concat}([\textbf{0}, \textbf{255}], \text{dim}=0)
\end{equation}
We next provide a detailed description of the key components used in our model, including the Font Size-Aware Modulation module, the Small-Character Aware Augmentation strategy, and the Task-Specific Region Weighted Loss employed during training.

\begin{figure}[!t]
\centering
\includegraphics[width=\textwidth]{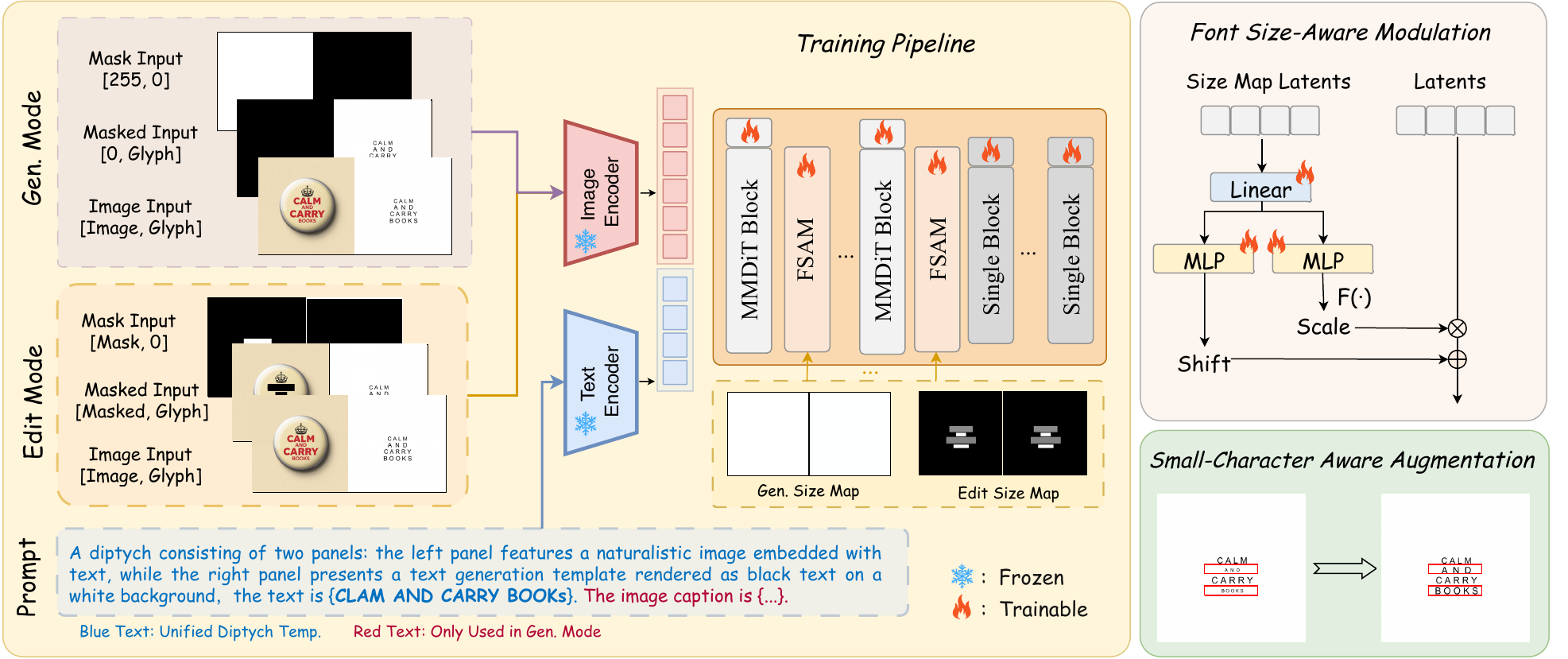}
\caption{\textbf{Overview of InnoText.} The model adopts different input paradigms and size maps for the editing and generation modes. In our framework, we incorporate a Font Size-Aware Modulation (FSAM) module that dynamically enhances the model’s attention to text regions of varying font sizes. In addition, we adopt a Small-Character Aware Augmentation strategy, which selectively enlarges small-font text samples to improve the generation quality of fine-scale characters.}
\label{fig:framework}
\vspace{-10pt}
\end{figure}

\subsection{Font Size-Aware Modulation}
Font Size-Aware Modulation (FSAM) enhances the model’s sensitivity to text regions of varying font sizes through an adaptive nonlinear feature calibration mechanism. Unlike conventional normalization-based modulation that applies uniform affine transformations, FSAM dynamically adjusts scaling and shifting parameters conditioned on font-size cues, improving fine-grained representation across multiple text scales.

The core of FSAM lies in the size map latents, which encode spatially varying font-size priors into the feature space. Unlike a binary mask, the size map is a grayscale image that assigns distinct intensity values to each detected text region, where the gray level indicates the degree of attention corresponding to its font size. To construct the map, we first estimate the approximate font height $F_h$ for each detected character region as:
\begin{equation}
F_h = \max\big(\min(h_{\text{bbox}}, w_{\text{bbox}}), 1\big),
\end{equation}
where $h_{\text{bbox}} = \max(y) - \min(y)$ and $w_{\text{bbox}} = \max(x) - \min(x)$ denote the height and width of the region’s bounding contour, respectively. To emphasize small-text regions, we compute the inverse font height and assign it to the corresponding spatial positions in the size map. The resulting grayscale map captures local scale variations across the text layout and is normalized to the range $[0,1]$ for inter-sample consistency.

As illustrated in Fig.~\ref{fig:framework}, this size map is patchfied and reshaped into a token sequence $S$, which is linearly projected to obtain latent embeddings $S'$ with the same dimensionality as the hidden states. These embeddings serve as size map latents, enabling cross-scale conditioning of hidden representations. Two parallel MLP branches then decode $S'$ into modulation parameters for scale and shift operations, respectively. 


The scale modulation $\gamma(s)$ is generated in two steps: the scale MLP($\text{MLP}_{\text{shift}}$) produces bounded outputs $s \in [0,1]$ via a sigmoid activation, which are subsequently transformed by a nonlinear amplification function:

\begin{equation}
\gamma(s) = 1 + (\gamma_{\text{max}} - 1) \cdot \sigma \big(k \cdot (s - 0.5)\big)
\end{equation}

\noindent where $\sigma$ is the Sigmoid function, $k$ controls the sharpness of the response, and $\gamma_{max}$ defines the maximum amplification ratio.

The shift modulation is generated by a separate MLP with a Tanh activation, scaled by a predefined parameter $\lambda_{\text{shift}}$ that controls the modulation amplitude:
\begin{equation}
\beta(S') = \lambda_{\text{shift}} \cdot \tanh(\text{MLP}_{\text{shift}}(S'))
\end{equation}
This scaling parameter $\lambda_{\text{shift}}$ provides flexible control over the intensity of the shift modulation, enabling more stable feature calibration across different text scales.

The final modulated hidden state is then computed as:
\begin{equation}
H' = H \cdot \gamma(s) + \beta(S')
\end{equation}
\noindent where $H$ represents the hidden states from the DiT block.


By explicitly coupling the hidden activations with size-aware latent modulation, FSAM allows the network to adaptively recalibrate its internal representations based on font-scale priors. This design effectively strengthens the attention to small-text regions while maintaining visual consistency across scales, thus enhancing the model’s capability in both text editing and generation tasks.

\subsection{Small-Character Aware Augmentation}

To bolster the model's structural fidelity and recognition performance for diminutive text, we introduce the SCAA strategy. This approach adaptively recalibrates the model's focus toward fine-grained, intricate details via stochastic regional magnification while maintaining global spatial coherence.

Specifically, we categorize a character instance as small-scale if its estimated font height $h$ falls below a predefined geometric threshold $\tau$. For these identified regions, we execute a probabilistic spatial scaling defined by a base factor $\lambda$ and a random perturbation $\delta$:

\begin{equation}
R' = \text{Resize}(R, \lambda + \delta), \quad \delta \sim \mathcal{U}(-\epsilon, \epsilon),
\end{equation}

where $R$ denotes the original region and $\mathcal{U}(-\epsilon, \epsilon)$ introduces controlled variability. This stochasticity serves as a form of foveal augmentation, enriching the model's exposure to diverse resolution scales and mitigating representation degradation in micro-scale text.

During inference, this scaling is symmetrically applied to editing tasks to ensure train-test consistency, thereby enhancing the fidelity of localized edits. Conversely, for text generation, we bypass this scaling to preserve the original font-size distribution of the input reference, ensuring a seamless and natural visual output.




\subsection{Task-Specific Region Weighted Loss}

During training, we employ the flow matching loss as the primary optimization objective to guide the model in generating predictions that are consistent with the ground-truth images under conditional inputs. Since our model simultaneously addresses both editing and generation tasks, we introduce task-specific loss formulations tailored to different input configurations, namely the combinations of $I_{input}$ and $I_{mask}$. This design enables effective unification and cooperative optimization across tasks.


\begin{equation}
\mathcal{L}_{\text{total}} =
\begin{cases}
\| S \cdot \mathcal{L}_{\text{FM}} \|^2_2, & \text{if editing task} \\
\mathcal{L}_{\text{FM}}, & \text{if generation task}
\end{cases}
\end{equation}

For the editing task, where only partial text regions require modification, we apply a region-weighted loss guided by a font size map. Higher weights are assigned to masked or modified areas to emphasize precise local refinement. In contrast, for the generation task, where the model synthesizes full-text images from scratch, we adopt a global loss to supervise the entire output, ensuring holistic structural consistency and overall visual fidelity.

This unified yet differentiated loss strategy aligns the optimization objectives with the specific requirements of each task, enhancing the model’s robustness and performance across diverse visual text scenarios.
\section{Experiments}

\begin{table}[t]
\centering
\caption{Quantitative comparison of editing and generating tasks, the best results are \textbf{bolded}. More results can be found in the supplementary materials.}
\begin{tabular}{c|c|ccc|ccc}
\toprule[1.1pt]
\multirow{2}{*}{\textbf{Tasks}} & \multirow{2}{*}{\textbf{Methods}} & \multicolumn{3}{c|}{\textbf{English}}                        & \multicolumn{3}{c}{\textbf{Chinese}}                         \\ \cmidrule(lr){3-8}
                      &                          & \textbf{Sen. Acc $\uparrow$}        & \textbf{NED $\uparrow$}            & \textbf{LPIPS $\downarrow$}          & \textbf{Sen. Acc $\uparrow$}       & \textbf{NED $\uparrow$}            & \textbf{LPIPS $\downarrow$}          \\ \midrule
\multirow{5}{*}{Edit} & Flux-Fill~\cite{flux2024}                 & 0.1342          & 0.2481          & 0.1650          & 0.0191          & 0.0443          & 0.1276          \\
                      & Anytext~\cite{tuo2023anytext}                  & 0.6839          & 0.8632          & 0.1234          & 0.6410          & 0.8209          & 0.1093          \\
                      & Anytext2~\cite{tuo2024anytext2}                 & 0.7893          & \textbf{0.9082} & 0.1739          & 0.7079          & 0.8419          & 0.1509          \\
                      & TextFlux~\cite{xie2025textflux}                 & 0.7732          & 0.8908          & 0.0838          & 0.7164          & 0.8498          & \textbf{0.0567} \\
                      & \textbf{Ours}                & \textbf{0.7988} & 0.9016          & \textbf{0.0786} & \textbf{0.7257} & \textbf{0.8579} & 0.0591          \\ \midrule
\multirow{5}{*}{Gen.} & Flux-Fill~\cite{flux2024}                 & 0.0126          & 0.0635          & 0.6232          & 0.0057          & 0.0200          & 0.6402          \\
                      & Anytext~\cite{tuo2023anytext}                   & 0.4707          & 0.7509          & 0.6463          & 0.4584          & 0.6440          & 0.6575          \\
                      & Anytext2~\cite{tuo2024anytext2}                 & 0.5638          & 0.7712          & 0.6330          & 0.5262          & 0.6636          & 0.5979          \\
                      & TextFlux~\cite{xie2025textflux}                 & 0.0071          & 0.0623          & 0.5968          & 0.0187          & 0.0470          & 0.6527          \\
                      & \textbf{Ours}                 & \textbf{0.6586} & \textbf{0.8066} & \textbf{0.4787} & \textbf{0.5907} & \textbf{0.6837} & \textbf{0.5311} \\ \bottomrule
\end{tabular}
\label{tab:edit_gen}
\end{table}

\subsection{Experiments Setting}
\noindent\textbf{Implementation Details.}
Our base model is Flux-Fill~\cite{flux2024}. During training, we alternate between the editing and generation tasks with an equal probability of 0.5. We use a batch size of 2 and optimize the model using the Prodigy optimizer with a weight decay of 0.01. The LoRA rank is set to 128. Training is conducted on 8 NVIDIA A100 GPUs for a total of 30,000 steps. For small-character aware augmentation, the size threshold $\lambda$ and scaling factor $\epsilon$ are set to 1.5 and 0.3, respectively. All images used in the experiments were resized to 512×512 pixels. 

\noindent\textbf{Evaluation Metrics.}
Following previous works, we evaluate both textual fidelity and perceptual quality for visual text editing and generation tasks. The evaluation metrics include Sentence Accuracy (Sen. Acc), which measures exact text correctness, Normalized Edit Distance (NED), which reflects character-level similarity between predictions and ground truth, and Learned Perceptual Image Patch Similarity (LPIPS), which assesses perceptual visual quality. The AnyText-Benchmark is used to evaluate editing performance, while InnoText-Bench is used for generation. Together, these metrics comprehensively measure the model’s ability to preserve textual accuracy and visual realism.

\noindent\textbf{Comparative Methods.}
We compare our method with several representative baselines covering both UNet-based and DiT-based diffusion architectures. The UNet-based methods include AnyText~\cite{tuo2023anytext} and AnyText2~\cite{tuo2024anytext2}, which support both visual text editing and generation within a unified framework. The DiT-based methods include Flux-Fill~\cite{flux2024} and TextFlux~\cite{xie2025textflux}, both designed for text editing tasks. All comparative methods are evaluated using their officially released pretrained weights.

\begin{figure*}[t]
\centering
\includegraphics[width=\textwidth]{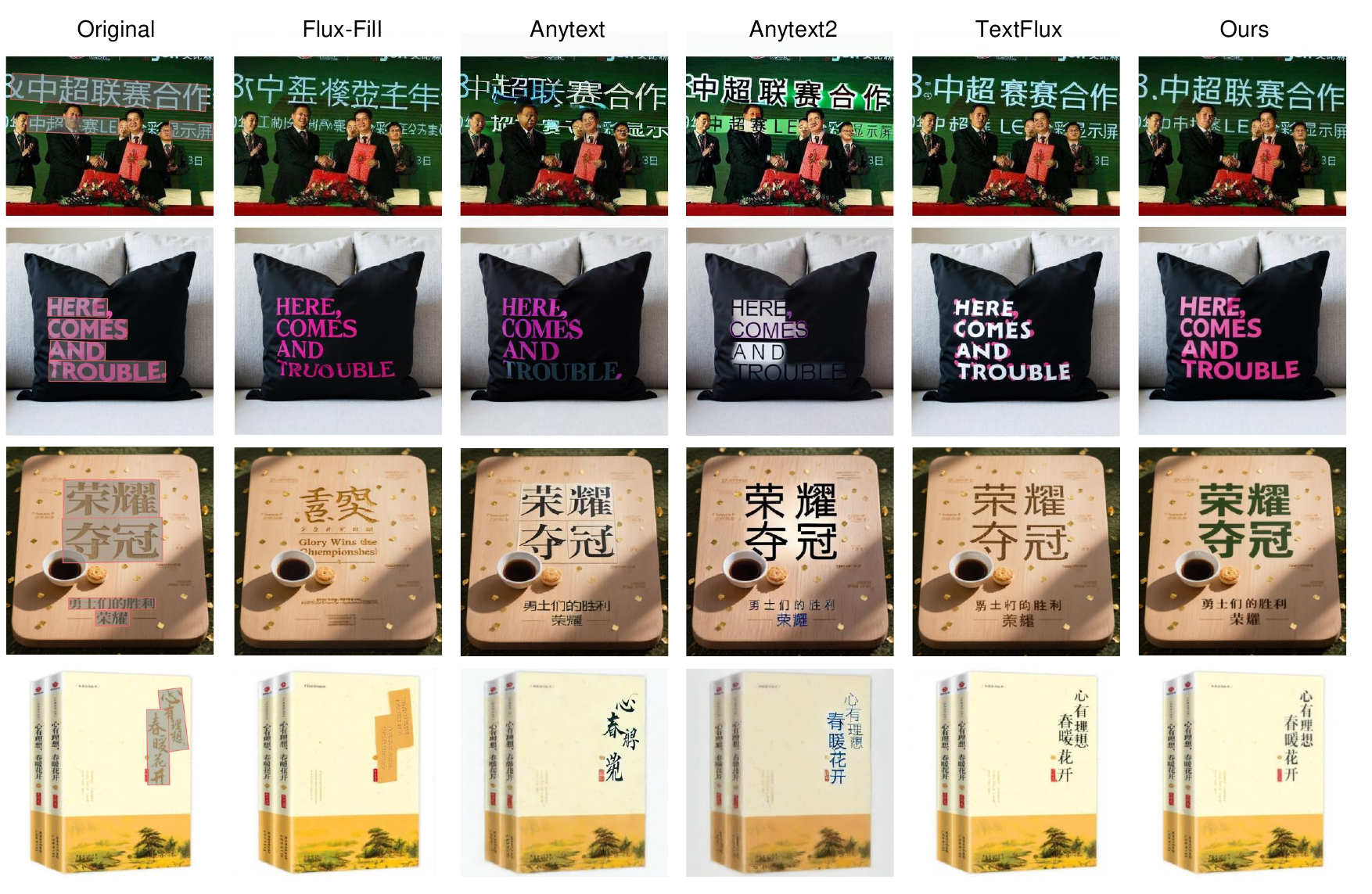}
\caption{\textbf{Qualitative comparison on editing tasks.} Our method generates text that blends seamlessly with complex backgrounds or non-flat background(first and second rows), maintains high visual quality across varying font sizes (third row), and produces coherent results under diverse text layouts, such as vertically arranged text (fourth row). More results can be found in the supplementary materials.}
\label{fig:comp_edit}
\end{figure*}


\begin{figure}[t]
\centering
\includegraphics[width=0.85\textwidth]{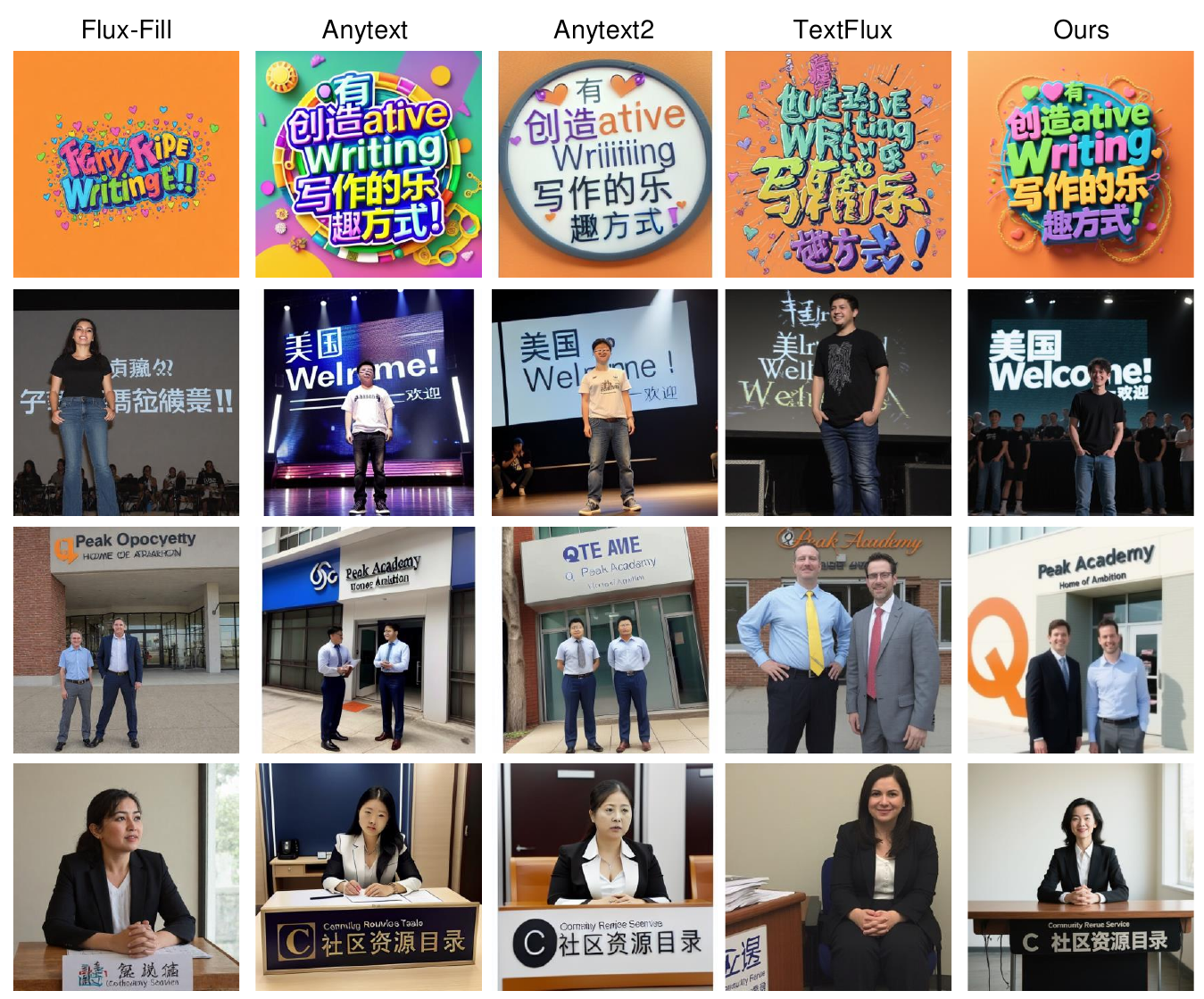}
\caption{\textbf{Qualitative comparison on generation task.} The visual text images generated by our method exhibit high aesthetic quality (first row), demonstrate appropriate integration with the background and higher background quality (second row), and achieve relative satisfactory results even for small-sized text (third and fourth rows). The corresponding prompt is provided in the supplementary materials.}
\label{fig:comp_gen}
\end{figure}

\begin{table*}[t!]
\centering
\caption{Ablation studies on visual text generation and editing, the best results are \textbf{bolded}.}
\begin{tabular}{l|ccc|ccc}
\toprule[1.1pt]
\multirow{2}{*}{\textbf{Methods}} & \multicolumn{3}{c|}{\textbf{Edit}}                        & \multicolumn{3}{c}{\textbf{Generation}}                         \\ \cmidrule(lr){2-7}
                         & \textbf{Sen. Acc $\uparrow$}       & \textbf{NED $\uparrow$}            & \textbf{LPIPS $\downarrow$}          & \textbf{Sen. Acc $\uparrow$}       & \textbf{NED $\uparrow$}            & \textbf{LPIPS $\downarrow$}          \\ \midrule
Anytext-FT                 & 0.5794          & 0.7138          & 0.1009          & 0.4622          & 0.6513          & 0.6186  \\
Anytext2-FT                 & 0.5882          & 0.7341          & 0.0532          & 0.5358          & 0.6801          & 0.5539  \\ \midrule
w/o FSAM                 & 0.5672          & 0.7048          & 0.0604          & 0.5508          & 0.6479          & 0.5928          \\
w/o SCAA                 & 0.5963          & 0.7196          & 0.0497          & 0.5615          & 0.6418          & 0.5539          \\
w/o $\mathcal{L}_{TSRW} $           & 0.6180          & \textbf{0.7533} & 0.0509          & 0.5892          & 0.6592          & 0.5671          \\
w/ all                   & \textbf{0.6374} & 0.7528          & \textbf{0.0484} & \textbf{0.5907} & \textbf{0.6837} & \textbf{0.5311} \\ \bottomrule
\end{tabular}
\label{tab:ablation}
\vspace{-10pt}
\end{table*}

\subsection{Quantitative Results}

As shown in Table~\ref{tab:edit_gen}, InnoText demonstrates consistently strong performance in visual text editing, ranking among the top methods across all evaluation metrics in both English and Chinese. It achieves the highest sentence accuracy and the lowest LPIPS, indicating its superior capability to preserve textual semantics and maintain high visual fidelity. Compared with representative baselines such as TextFlux and AnyText2, InnoText delivers more precise text reconstruction and better structural consistency, particularly in challenging Chinese text scenarios.

For visual text generation, InnoText likewise surpasses all competing methods, attaining the best sentence accuracy and the lowest LPIPS scores. These results highlight its strong ability to synthesize text that is both semantically accurate and visually coherent, confirming the effectiveness of our unified framework in handling diverse generation and editing tasks with consistent quality.

\subsection{Qualitative Results}

We conducted qualitative analyses to validate the effectiveness of our method in both tasks. As shown in Fig.~\ref{fig:comp_edit}, Flux-Fill fails to generate Chinese text due to its lack of Chinese encoding capability. Although AnyText and AnyText2 can produce Chinese characters, their generation quality noticeably deteriorates in complex scenes or when handling text of varying sizes. TextFlux demonstrates limited fidelity in character details, with certain strokes appearing distorted or incorrectly formed.

As shown in Fig.~\ref{fig:comp_gen}, for visual text generation, our method produces images with superior background quality and text rendering fidelity compared with competing approaches. The generated text not only maintains high semantic accuracy, but also exhibits enhanced visual aesthetics, achieving a more natural and visually pleasing integration between text and background.

\subsection{Ablation Studies}

The results of our ablation studies are shown in Table~\ref{tab:ablation}. We evaluate the performance of AnyText and AnyText2 after fine-tuning on the InnoText-30K dataset, and further analyze the impact of each module through ablation studies. 
Previous methods such as AnyText and AnyText2 show performance improvements after fine-tuning on the InnoText-30K dataset. However, their results still remain inferior to our method.

Among all modules, the Font Size-Aware Modulation (FSAM) contributes the most, as its removal increases the LPIPS score from 0.5311 to 0.5928, particularly degrading generation quality. Excluding the Small Character Aware Augmentation consistently lowers sentence accuracy and visual similarity, especially in editing, underscoring its importance in enhancing small-text clarity. Removing the Task-Specific Region Weighted Loss leads to moderate performance degradation, including a 1.93\% drop in editing accuracy, indicating its role in task-aware optimization. The best overall performance is achieved when all components are combined, demonstrating their complementary contributions to both visual quality and text generation accuracy.

\begin{figure}[t]
\centering
\includegraphics[width=\textwidth]{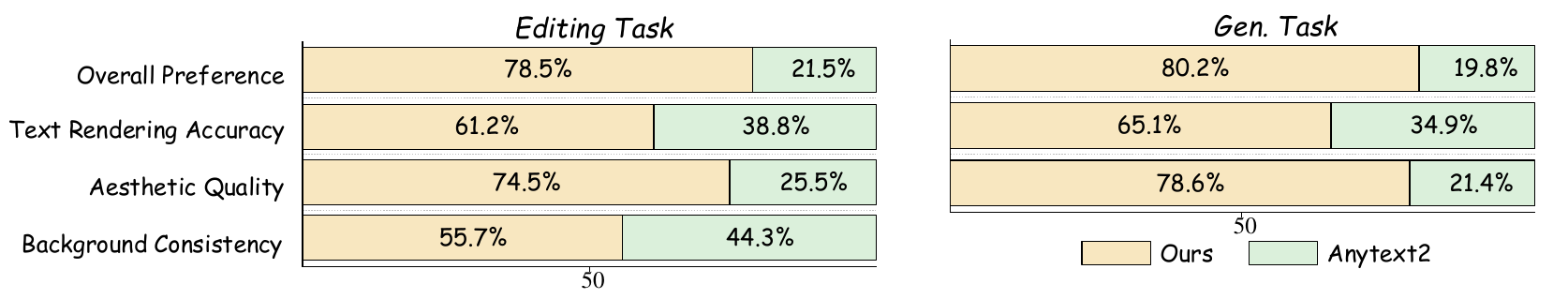}
\caption{Results of Human Perceptual Evaluation.}
\label{fig:user_study}
\end{figure}

\subsection{Human Evaluation}

To rigorously assess the perceptual superiority of InnoText, we conducted a comprehensive double-blind Side-by-Side (SbS) user study involving 15 participants with diverse backgrounds. We curated a balanced evaluation set comprising 100 cases (50 for generation and 50 for editing) and compared our model against the state-of-the-art AnyText2. Respondents were tasked with evaluating paired outputs based on four critical dimensions: text rendering accuracy (structural integrity and OCR accuracy), aesthetic quality (visual realism and texture quality), background consistency (seamless integration between text and background, particularly for editing), and overall preference. To ensure statistical objectivity, the order of images was randomized, and participants were blinded to the underlying model identities.

As illustrated in Fig.~\ref{fig:user_study}, InnoText consistently achieves a dominant lead over the baseline across both task categories, with its overall preference rate exceeding 78\%. Quantitative analysis of the feedback indicates that our DiT-based architecture effectively resolves the low-level precision vs. high-level harmony trade-off, which often plagues UNet-based frameworks. Notably, in Chinese text scenarios, our model was frequently cited for its superior handling of intricate stroke connectivity and font-style consistency, while the competitors often exhibited blurred contours or structural artifacts. These results underscore the robustness of InnoText in generating high-fidelity visual text that aligns with human perceptual expectations in complex real-world environments.

\subsection{Failure Cases Studies}
\begin{figure}[t]
\centering
\includegraphics[width=0.9\textwidth]{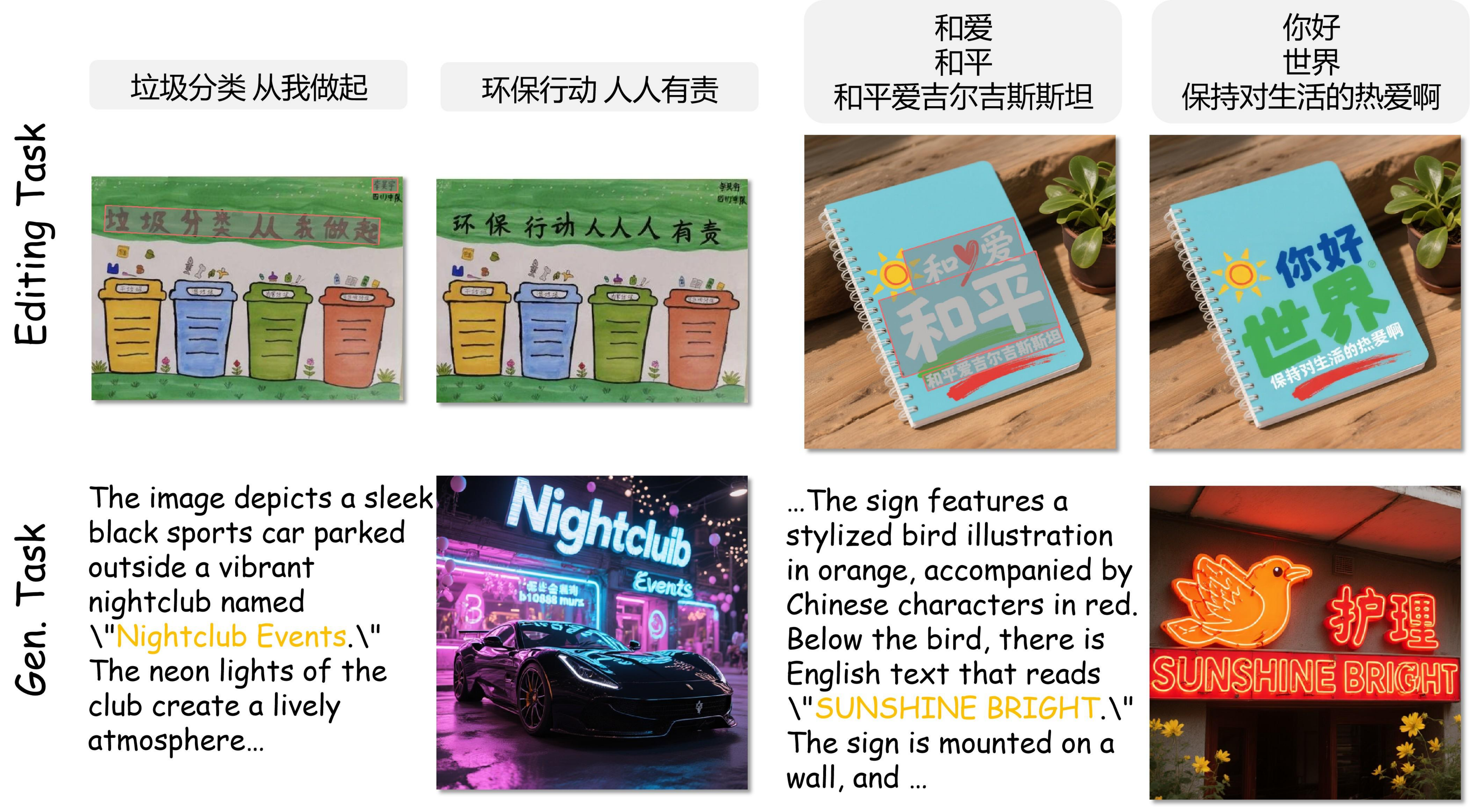}
\caption{Representative failure cases.}
\label{fig:fail_case}
\end{figure}

We analyze representative failure cases, as illustrated in Fig.~\ref{fig:fail_case}. For the editing task, overly large mask regions may occasionally result in local text duplication, likely due to ambiguous spatial constraints when the editable area covers extensive contextual content. For small and densely structured characters, minor stroke-level inaccuracies can still appear, especially under extreme font scales. In the generation task, redundant characters and complex Chinese glyphs with high stroke density remain relatively challenging. Despite the font size-aware modulation and small-character aware augmentation improving multi-scale and fine-grained representations, precise structural modeling of repetitive patterns and intricate stroke layouts is not yet fully resolved.

These observations suggest that future work could further enhance spatial controllability and character-level structural consistency within unified DiT-based text rendering frameworks.

\section{Conclusion}

In this paper, we present InnoText, a unified Diffusion Transformer framework that harmonizes visual text generation and editing within a cohesive paradigm. By integrating Font Size-Aware Modulation (FSAM) and Small-Character Aware Augmentation (SCAA), our approach alleviates long-standing bottlenecks in micro-scale text representation and complex non-Latin script modeling. We further introduce a Task-Specific Region Weighted Loss to balance localized precision with global semantic coherence. Coupled with the InnoText-30K datasets, a high-fidelity bilingual dataset, our work establishes a rigorous benchmark for visual text research. Extensive quantitative and qualitative evaluations, including a blind user study, demonstrate that InnoText consistently achieves state-of-the-art performance across diverse scenarios. Ultimately, this work not only offers a robust solution for high-precision typographic synthesis but also provides significant insights into the development of scale-aware and multi-modal foundational models for real-world creative applications.

\bibliographystyle{splncs04}
\bibliography{main}

\clearpage
\setcounter{page}{1}

\setcounter{section}{0}
\section{More Quantitative Results}

In addition to evaluating the visual text editing capabilities of various methods using the AnyText-benchmark, we further conducted experiments on the test set of our proposed dataset. The quantitative results are summarized in the Table below.

Our method, InnoText, consistently outperforms all baselines on both languages. Specifically, InnoText achieves the highest Sentence Accuracy (English: 0.6894, Chinese: 0.6374) and the lowest LPIPS score (English: 0.0507, Chinese: 0.0484), indicating superior semantic correctness and visual fidelity. Compared to the baseline TextFlux, InnoText shows noticeable improvements in both Sentence Accuracy and LPIPS, while also maintaining competitive NED values. These results demonstrate the effectiveness and generalizability of our method across different languages and evaluation metrics.

\begin{table}[h!]
\centering
\caption{Quantitative results of various methods on the editing task .}
\resizebox{0.9\textwidth}{!}{
\begin{tabular}{c|ccc|ccc}
\toprule[1.1pt]
\multirow{2}{*}{\textbf{Methods}} & \multicolumn{3}{c|}{\textbf{English}}                        & \multicolumn{3}{c}{\textbf{Chinese}}                         \\ 
\cmidrule(lr){2-7}
                         & \textbf{Sen. Acc $\uparrow$}        & \textbf{NED $\uparrow$}            & \textbf{LPIPS $\downarrow$}          & \textbf{Sen. Acc $\uparrow$}        & \textbf{NED $\uparrow$}            & \textbf{LPIPS $\downarrow$}          \\ 
\midrule
FluxFill                 & 0.2814          & 0.4105          & 0.1056          & 0.0517          & 0.0866          & 0.1022          \\
Anytext                  & 0.5813          & 0.7633          & 0.1098          & 0.5578          & 0.6929          & 0.1276          \\
Anytext2                 & 0.6603          & 0.7902          & 0.1224          & 0.6213          & 0.7349          & 0.1217          \\
TextFlux                 & 0.6490          & 0.7987          & 0.0616          & 0.6120          & 0.7302          & 0.0628          \\
\textbf{Ours}           & \textbf{0.6894} & \textbf{0.8356} & \textbf{0.0507} & \textbf{0.6374} & \textbf{0.7528} & \textbf{0.0484} \\ 
\bottomrule
\end{tabular}
}
\label{tab:editing_task_results}
\end{table}


Following Flux-Text, we extended our evaluation using 5,000 random samples from MARIO-Eval. As shown in the table, our model trained on InnoText-30K attains the best overall performance across nearly all metrics, when trained with AnyWord-3M, our framework achieves highly competitive results.

\begin{table}[h!]
\centering
\caption{Quantitative results of various methods using the Mario-Eval benchmark.}
\resizebox{0.9\textwidth}{!}{
\begin{tabular}{c|cccc}
\toprule[1.1pt]
\textbf{Methods}   & \textbf{FID$\downarrow$}  & \textbf{CLIPScore$\uparrow$} & \textbf{OCR-Accuracy$\uparrow$} & \textbf{OCR-F1$\uparrow$} \\ \midrule
Anytext2  & 17.47         & 0.3215             & 0.5613                 & 0.7853          \\
TextFlux  & 5.25          & 0.3287             & 0.6121                  & 0.8023          \\
Flux-Text & 4.78          & \underline{0.3319}              & 0.6312                    & 0.8455 \\
\textbf{Ours (30K)}      & \underline{3.84} & \textbf{0.3356}   & \textbf{0.6501}         & \textbf{0.8605}          \\ 
\textbf{Ours (3M)}      & \textbf{3.76 }& 0.3318       & \underline{0.6471}         & \underline{0.8583}      \\ \bottomrule[1.1pt]
\end{tabular}
}
\end{table}

In addition, we also explored the impact of different LoRA ranks on text generation performance, with the results shown in the Table~\ref{tab:lora_rank_impact}.

\begin{table}[h!]
\centering
\caption{The impact of different LoRA ranks on the results}
\resizebox{0.9\textwidth}{!}{
\begin{tabular}{c|ccc|ccc}
\toprule[1.1pt]
\multirow{2}{*}{\textbf{Rank}} & \multicolumn{3}{c|}{\textbf{Edit}} & \multicolumn{3}{c}{\textbf{Generation}} \\ 
\cmidrule(lr){2-7}
                      & \textbf{Sen. Acc $\uparrow$}  & \textbf{NED $\uparrow$}     & \textbf{LPIPS $\downarrow$}  & \textbf{Sen. Acc $\uparrow$}  & \textbf{NED $\uparrow$}    & \textbf{LPIPS $\downarrow$}  \\ 
\midrule
32                    & 0.5902    & 0.7327  & 0.5451 & 0.5133    & 0.6219 & 0.5882 \\
64                    & 0.6307    & 0.7802  & 0.5004 & 0.5609    & 0.6651 & 0.5537 \\
128                   & 0.6586    & 0.8066  & 0.4787 & 0.5907    & 0.6837 & 0.5311 \\
256                   & 0.6693    & 0.8154  & 0.4612 & 0.5889    & 0.6902 & 0.5198 \\ 
\bottomrule
\end{tabular}
}
\label{tab:lora_rank_impact}
\end{table}


Furthermore, we evaluated our model against the leading open-source SOTAs, Flux-Text for editing and EasyText for generation. While our model yields competitive results using the high-quality InnoText-30K dataset, it achieves state-of-the-art performance particularly in Chinese tasks. When trained on the larger AnyWord-3M (consistent with Flux-Text), our framework outperforms Flux-Text across nearly all metrics. The result demonstrates the superiority of our architecture and joint training.

\begin{table}[h!]
\centering
\caption{Comparison with open-source SOTAs}
\resizebox{\textwidth}{!}{
\begin{tabular}{c|c|c|ccc|ccc}
\toprule[1.1pt]
\multirow{2}{*}{\textbf{Tasks}} & \multirow{2}{*}{\textbf{Methods}} & \multirow{2}{*}{\textbf{GPU Hrs $\downarrow$}} & \multicolumn{3}{c|}{\textbf{English}}                        & \multicolumn{3}{c}{\textbf{Chinese}}                         \\ 
\cmidrule(lr){4-9}
                      &                          &                          & \textbf{Sen. Acc $\uparrow$}        & \textbf{NED $\uparrow$}            & \textbf{LPIPS $\downarrow$}          & \textbf{Sen. Acc $\uparrow$}       & \textbf{NED $\uparrow$}            & \textbf{LPIPS $\downarrow$}          \\ 
\midrule
\multirow{3}{*}{Edit} 
& Flux-Text (3M)                
& $\sim$960 
& \underline{0.8175} & \underline{0.9193} & \underline{0.0674} 
& 0.7213 & \underline{0.8555} & \underline{0.0487} \\
                      
& \textbf{Ours (30K)}                
& \textbf{$\sim$ 128} 
& 0.7988 & 0.9016 & 0.0786 
& \textbf{0.7257} & \textbf{0.8579} & 0.0591 \\ 

& \textbf{Ours (3M)}                
& \underline{$\sim$768} 
& \textbf{0.8224} & \textbf{0.9251} & \textbf{0.0638} 
& \underline{0.7236} & 0.8523 & \textbf{0.0472} \\ 

\midrule
\multirow{3}{*}{Gen.} 
& EasyText (1M)                
& - 
& 0.6513 & 0.7932 & 0.5301 
& 0.5619 & 0.6643 & \underline{0.5397} \\

& \textbf{Ours (30K)}                 
& $\sim$128 
& \underline{0.6586} & \underline{0.8066} & \textbf{0.4787} 
& \textbf{0.5907} & \textbf{0.6837} & \textbf{0.5311} \\ 

& \textbf{Ours (3M)}                
& $\sim$768 
& \textbf{0.6597} & \textbf{0.8119} & \underline{0.5032} 
& \underline{0.5803} & \underline{0.6798} & 0.5414 \\ 

\bottomrule
\end{tabular}
}
\label{tab:baseline_compare}
\end{table}

\begin{figure}[h]
\centering
\includegraphics[width=0.8\linewidth]{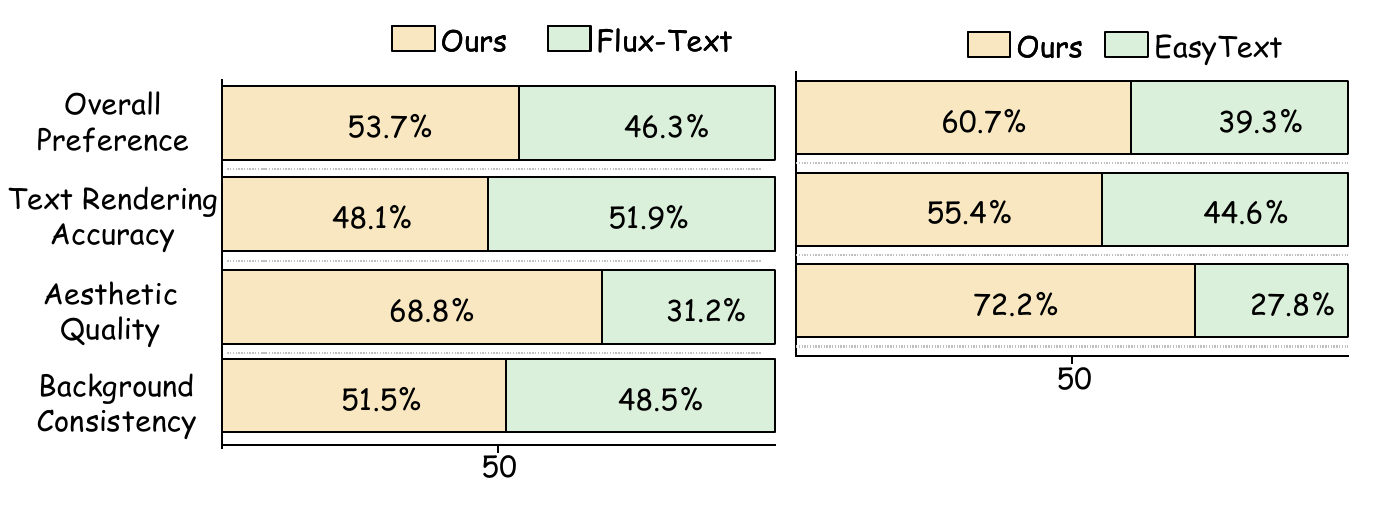}
\caption{More human study.}
\label{fig:human_study}
\end{figure}

As shown in Fig.~\ref{fig:human_study}, we strengthened the human study by increasing the number of participants to 45 and expanding the comparisons to include the strongest task-specific SOTA models: Flux-Text (editing) and EasyText (generation). As illustrated in the figure below, our model consistently achieves superior overall preference rates against these strong baselines.

\section{More Qualitative Results}

In this section, we present additional qualitative results for both the text editing and generation tasks. More examples of editing results are shown in Fig.~\ref{fig:comp_edit}, while additional generation results are illustrated in Fig.~\ref{fig:comp_gen}. These results further demonstrate that our proposed model is capable of accurately generating small text content while preserving high-quality background details. Moreover, the generated images exhibit strong artistic coherence and aesthetic appeal, highlighting the effectiveness of our method in producing visually pleasing and semantically correct outputs.

\begin{figure}[t]
\centering
\includegraphics[width=\linewidth]{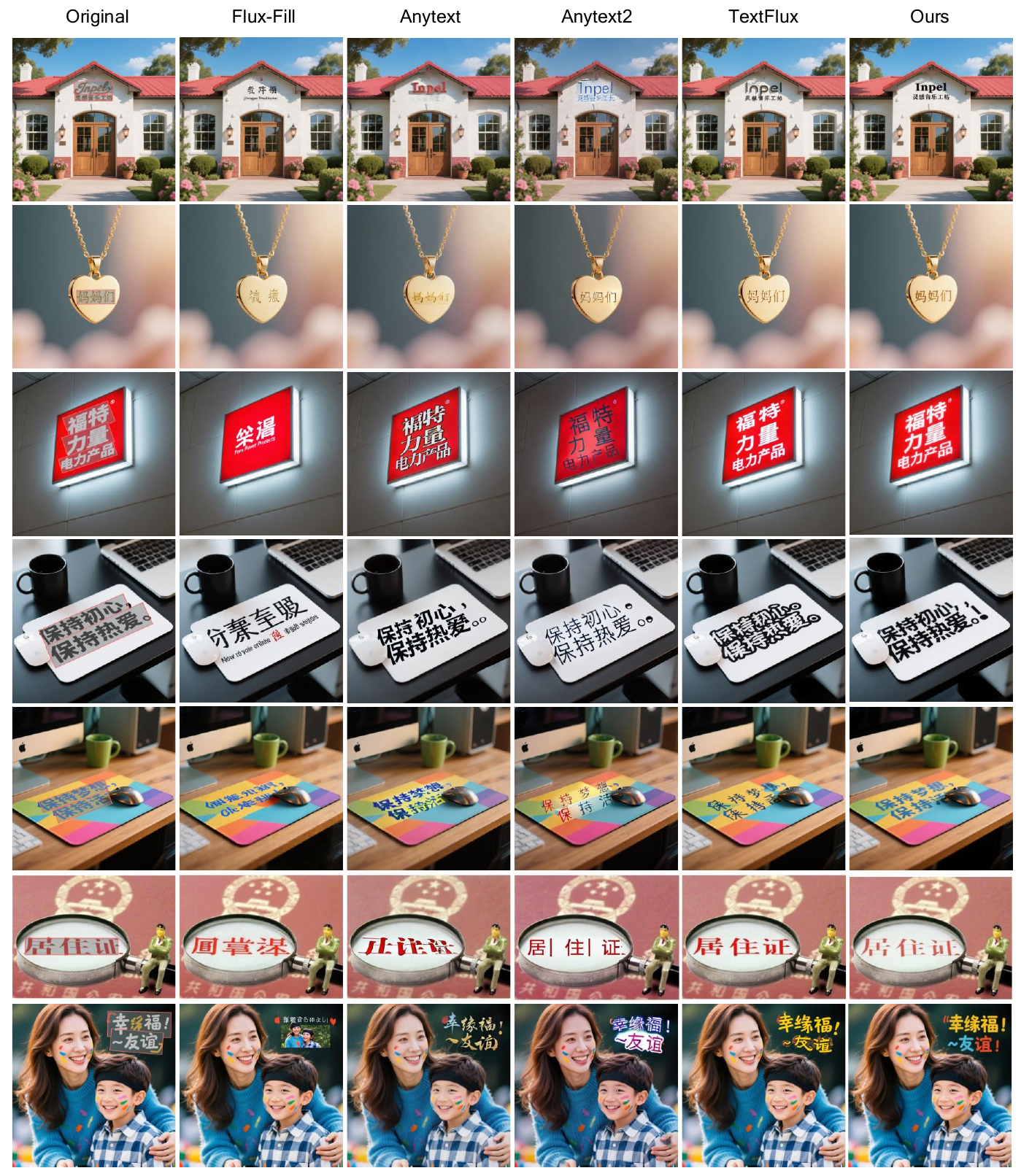}
\caption{Qualitative comparison on visual text editing task.}
\label{fig:comp_edit}
\end{figure}

\begin{figure*}[t]
\centering
\includegraphics[width=\linewidth]{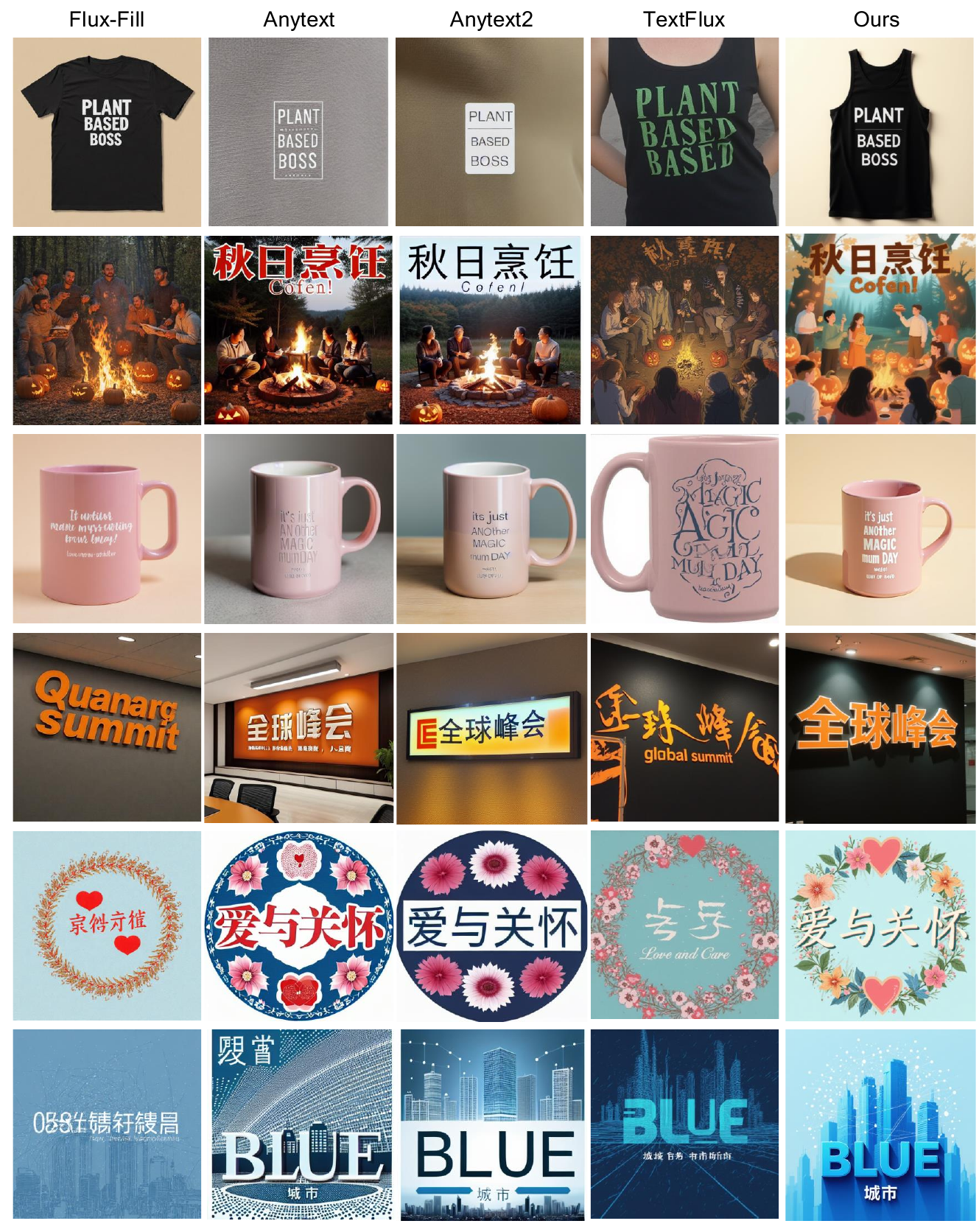}
\caption{Qualitative comparison on visual text generation task.}
\label{fig:comp_gen}
\end{figure*}

\section{The Prompts Used to Generate Images}
\begin{figure*}[!t]
\centering
\includegraphics[width=\linewidth]{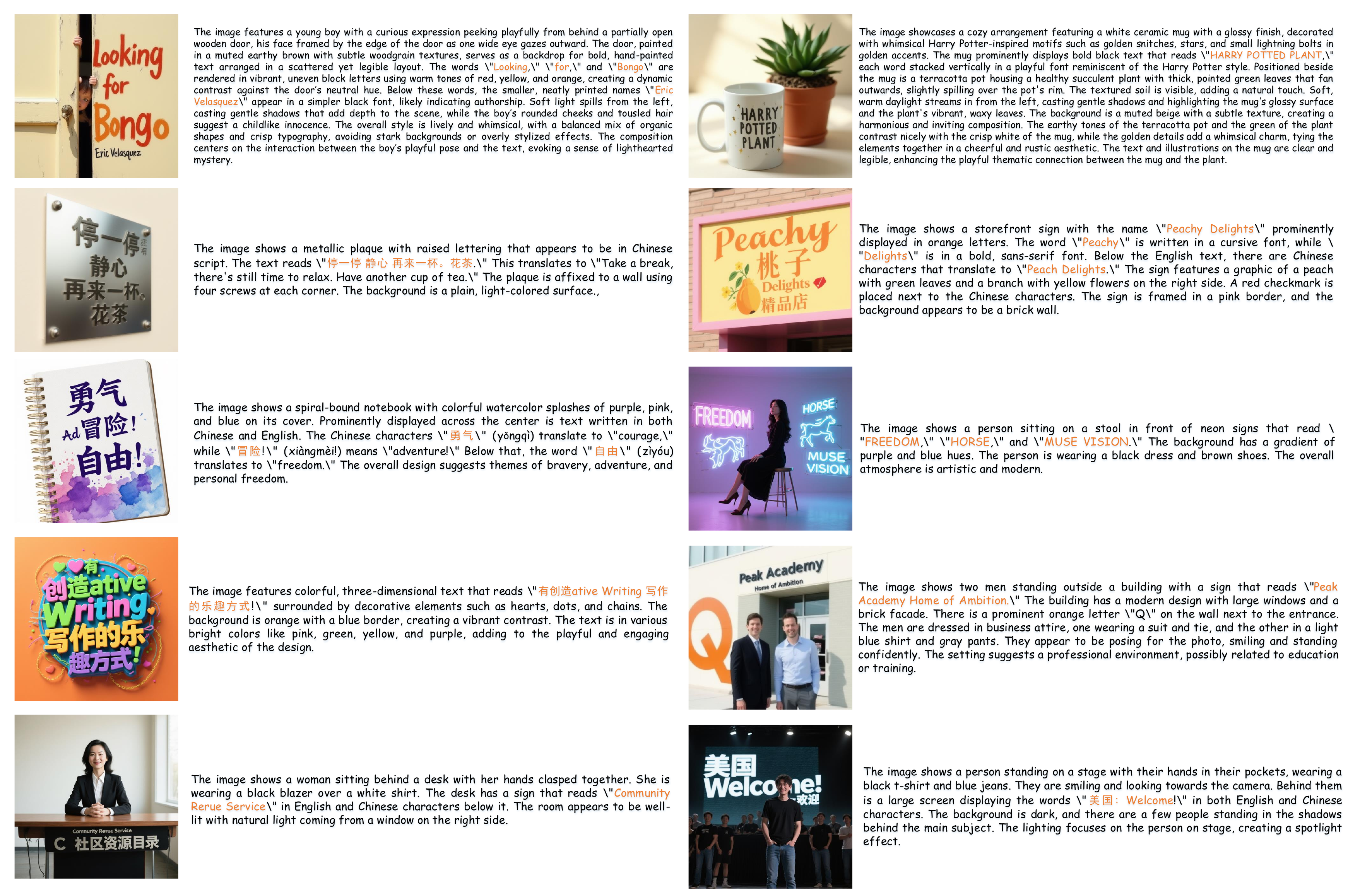}
\caption{The generated image and the corresponding caption content of the image.}
\label{fig:prompt_fig1}
\vspace{-10pt}
\end{figure*}

In our experiments, both visual text editing and generation tasks are performed using fixed prompt templates. For the editing task, the target text is inserted into the template, while for the visual text generation task, image captions are incorporated into the prompts to guide image synthesis. This section presents the specific captions used as prompts during the visual text generation experiments in the paper. The generated images along with their corresponding captions are shown in Fig.~\ref{fig:prompt_fig1}.

\section{Social Impact and Ethical Considerations}
Visual text generation and editing technologies represent a significant advancement in visual-language understanding, with applications in augmented reality, accessibility, digital media, and education. These methods enable dynamic manipulation of text within images, offering both practical and creative potential. As with many AI tools, their societal impact involves a complex balance of benefits and risks.
\subsection{Positive Effect}
On the positive side, this technology can enhance accessibility for visually impaired users, support multilingual communication, and enable personalized content in interactive environments. It also offers new possibilities in design, virtual experiences, and educational tools, fostering innovation across creative and technical domains.

\subsection{Nagetive Effect}
In contrast, the potential for misuse, such as generating deceptive content, forging visual evidence, or violating privacy, raises ethical concerns. Biases in training data may lead to unfair or culturally insensitive outputs, while a lack of regulation could allow harmful applications. Mitigation strategies such as watermarking, usage controls, and responsible deployment are essential to ensure safe and ethical use.

\end{document}